\documentclass{article}
\usepackage{graphicx}
\usepackage{amsmath,amssymb,amsfonts}
\usepackage{algorithmic}
\usepackage{textcomp}
\usepackage{booktabs}
\usepackage{xcolor}
\usepackage{caption}
\usepackage{subcaption}
\usepackage[hyperfootnotes=false]{hyperref}

\usepackage[final]{corl_2022} 

\title{Dominion: A New Frontier for AI Research}

\author{
  Danny Halawi\thanks{Equal Contribution} \hphantom{\tiny.}\thanks{University of California, Berkeley}\hphantom{a}, Aron Sarmasi\footnotemark[1] \hphantom{\tiny.}\thanks{Univeristy of California, Davis}\hphantom{a}, Siena Saltzen$^\ddag$, Joshua McCoy$^\ddag$  \\
}

\begin{document}
\maketitle


\begin{abstract}
    In recent years, machine learning approaches have made dramatic advances, reaching superhuman performance in Go, Atari, and poker variants. These games, and others before them, have served not only as a  testbed but have also helped to push the boundaries of AI research. Continuing this tradition, we examine the tabletop game Dominion and discuss the properties that make it well-suited to serve as a benchmark for the next generation of reinforcement learning (RL) algorithms. We also present the Dominion Online Dataset, a collection of over 2,000,000 games of Dominion played by experienced players on the Dominion Online webserver. Finally, we introduce an RL baseline bot that uses existing techniques to beat common heuristic-based bots, and shows competitive performance against the previously strongest bot, Provincial.
\end{abstract}



\section{Introduction}
	
    Games have long played a role in AI research, both as a test-bed, and as a moving goal-post, constantly driving innovation. From the heyday of chess agents, when Deep Blue beat Gary Kasparov, to more recent advances, like AlphaGo’s dark horse ascent to fame, games have both assisted AI research and provided something to aim for. As the AIs got better, the games they were applied to also got more complex. New game mechanics, such as the fog of war in StarCraft and the stochasticity of Poker, pushed researchers to adapt their methods to ever greater generality. In this paper, we argue that the deck-building strategy game Dominion \cite{Vaccarino2008} deserves to join the ranks of AI benchmark games, providing an RL-based bot in service of that benchmark.

Dominion has all of the abovementioned elements, but it also incorporates a mechanic that is not present in other popular RL benchmarks: every game is played with a different set of cards. Since each dominion card has a specific rule printed on it, and the set of 10 cards for a game are randomly picked from among hundreds of cards, no two games of Dominion can be approached the same way. Thus a key part of playing Dominion is adapting one’s inductive bias of how to play to the specific cards on the table. Although a general knowledge of which card combos are powerful is certainly helpful, with over $350 \choose 10$ possible kingdom card setups, it is insufficient to rely solely on inductive bias. In addition to its game mechanics, Dominion has an active player community, with regular tournaments, well documented strategy guides \cite{GameWiki}, and a popular online web server \cite{Meijer2017}. The size of the player community serves not only to increase the relevance and audience of Dominion related research, but it also increases the credibility of a bot who can beat the top players. 

In this paper, we lay out the reasons why Dominion is a strong benchmark candidate for AI systems, and we provide an RL player as a baseline for future research. We start by introducing the game and describing the various mechanics that are friendly to AI development. Next we describe a large dataset of 2,000,000 human-played game records, which we've already compiled 30,000 of them and have made easily available online. Finally, we introduce a deep RL based bot that can handily beat common heuristic based bots and and is competitive against the previous best AI, Provincial.


\section{Dominion} \label{sec:dominion}

\begin{figure}[ht!]
\centering
\begin{subfigure}{.3\textwidth}
  \centering
  \includegraphics[scale=0.1]{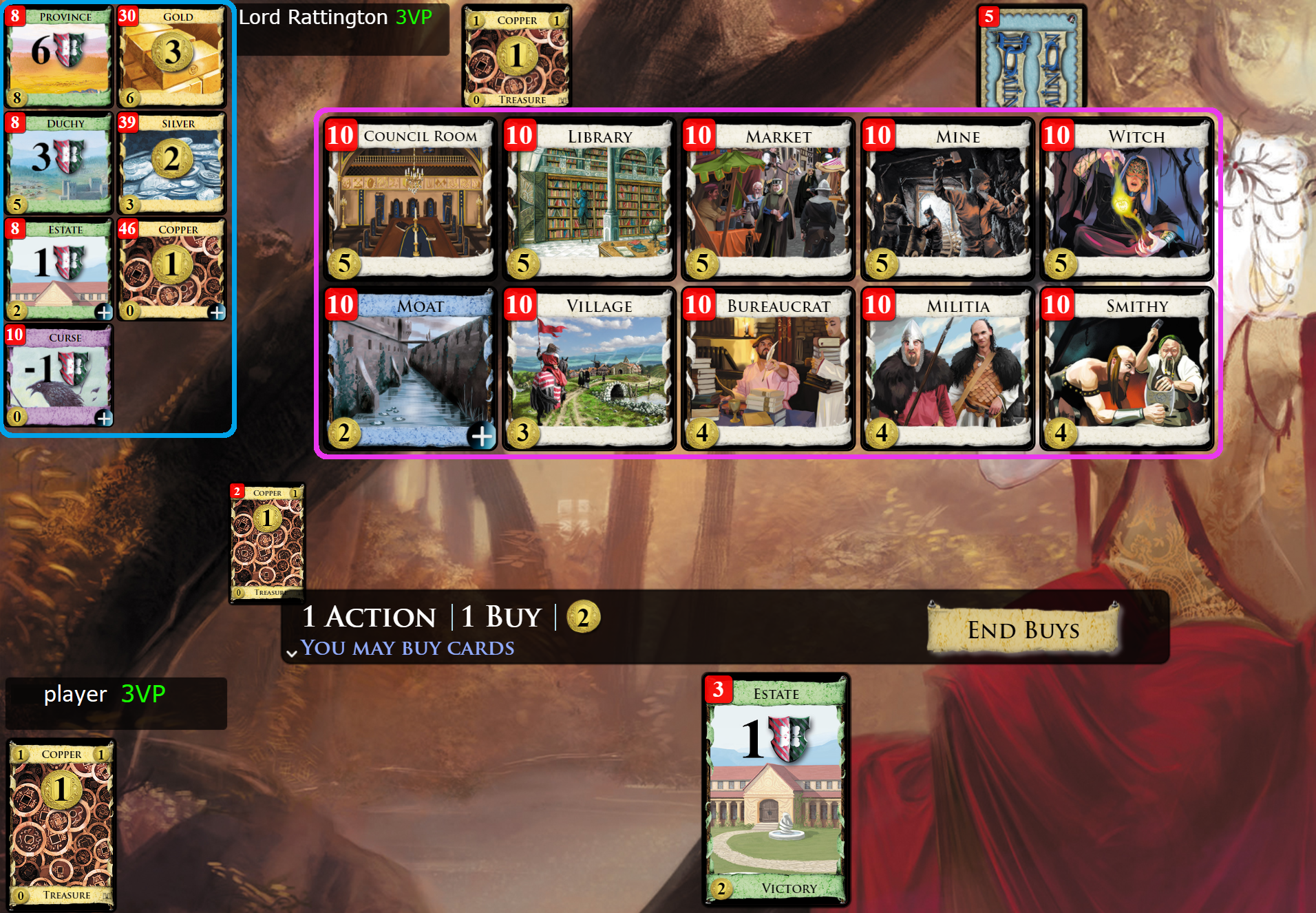}
\subcaption*{\hphantom{aaaa.}(a) Dominion Web Interface}
  \label{fig:main_a}
\end{subfigure}%
\begin{subfigure}{.8\textwidth}
  \centering
\includegraphics[scale=0.2]{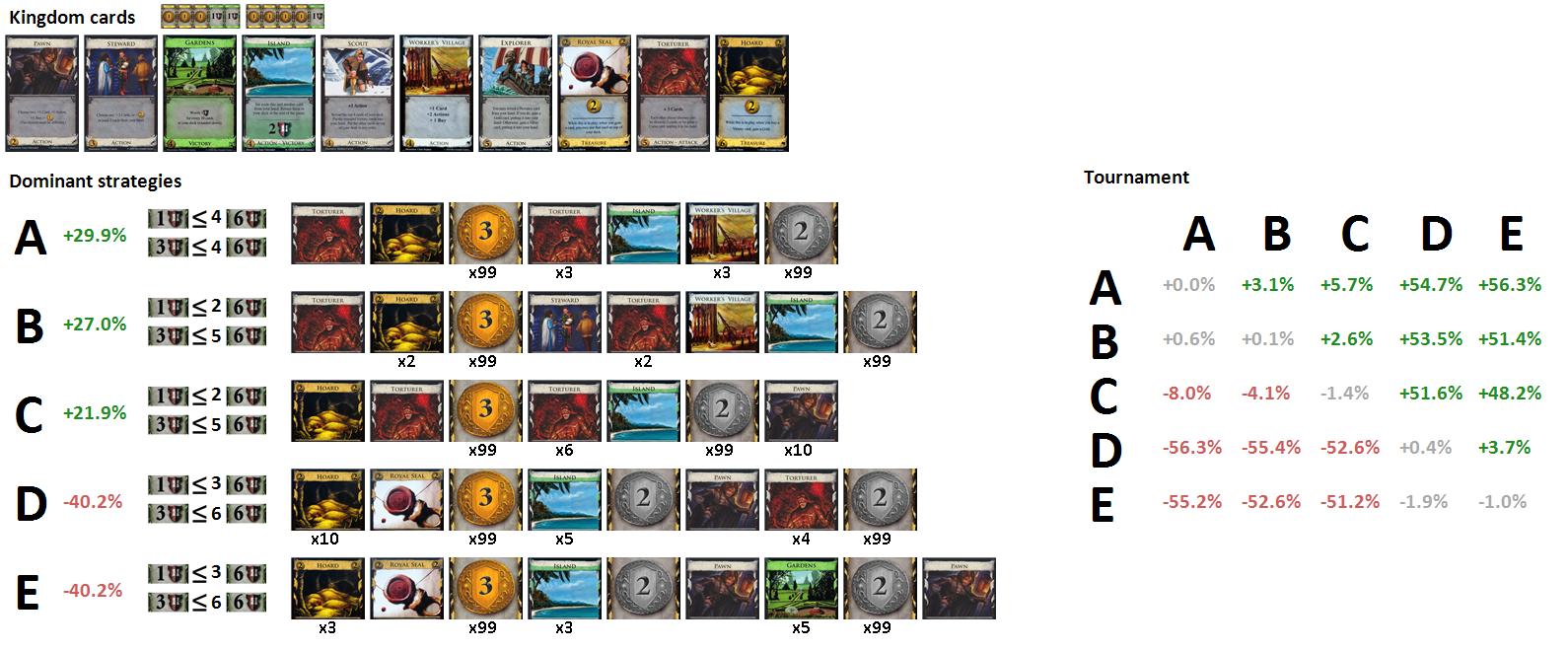}
\subcaption*{(b) Provincial Strategies}
\label{fig:main_b}
\end{subfigure}
\caption{\textbf{(a)} Typical game setup on the Dominion Online web server. The \textit{basic cards} in the blue rectangle are present in every game of Dominion, whereas the 10 \textit{kingdom cards} in the magenta rectangle are chosen randomly from a set of up to 350 cards. A dominion card has the card cost in the bottom left and the number of copies in the top left. \textbf{(b)} A visualization of the leading strategies Provincial discovered given a random cardset (including cards from later expansions). From left to right, we see 1) the expected win ratio when each of the five strategies plays against each other strategy averaged over 10,000 games, 2) the victory card purchase thresholds, and 3) the buy menus.}
\label{fig:main}
\end{figure}
Dominion is a turn-based, deck building strategy game for 2-6 players. To keep our project in scope, we have restricted ourselves to ranked 2-player games using the 2nd edition Dominion card set (commonly referred to as the \textit{base game}). The object of the game is to own the most \textit{victory points} (VPs) when the game ends, which typically means buying \textit{victory cards} over the course of the game. When the Province victory card pile runs out, or any three piles run out, the game ends. In a turn, a player may play a single action card and any number of treasure cards, and then buy a single card, before discarding the rest of their hand and everything they played that turn and drawing 5 cards from their deck to form the next turn’s hand. As cards are bought, they are placed directly in the discard pile, and are only drawn sometime later, i.e. once the deck runs out and the discard pile is shuffled to create a fresh deck. 


The vast majority of the complexity of Dominion is due not to its basic rules, described above, but rather to the large variety of cards that are available to play. During game setup, players randomly choose 10 different kingdom cards from among over 350 different available cards; 10 copies of each kingdom card is used to form the supply piles. In addition to the kingdom cards, every game is played with a set of base cards, as shown in Fig. \ref{fig:main}(a). This represents a lower bound of ${350 \choose 10} = 6 \cdot 10^{18}$ different possible game setups. Since each kingdom card has a unique rule printed on it, the ruleset, and thus the optimal strategy, are going to be different each game.

In addition to the unique mechanics, Dominion has a few other properties that make it especially attractive to AI researchers. Unlike Atari games where each pixel is a dimension, the dominion state space is quite small. Each game is played with only 17-25 different cards, and these cards can only be in a few different positions: the supply, the player’s deck, the player's hand, and the player's discard (modulo a few other locations that only come into play with certain cards). Based on our analysis of Dominion game logs (see section \ref{sec:doda}), games usually consist of 15-40 turns, and a player usually has to make less than a few hundred decisions in a typical game. These properties allow researchers to solve a very interesting problem without the high computational resource requirements entailed by other baselines.

	

\section{Dominion Online Dataset} \label{sec:doda}

One of the most popular ways to play dominion, especially given the high rate of self-isolation in the past year, is on the web server Dominion Online. Fortunately for the research community, the developers save game logs for all the games played on the site (over 2,000,000). It is these logs that we have downloaded,
processed, and will now present.

\renewcommand{\arraystretch}{0.95}
\begin{table}[ht!]
  \centering
  \caption{Statistics of 30,000 rated games of 2-player Dominion using only 2nd edition base game cards, downloaded from the Dominion Online web server.}
    \begin{tabular}{lr}
    \toprule
    Statistic & \multicolumn{1}{l}{Value} \\
    \midrule
    Avg game length & $31$ \\
    Avg VP totals/player & $20$ \\
    Avg margin of victory [VPs] & $8$ \\
    Avg gain decisions/player & $19$ \\
    Avg card plays/player & $44$ \\
    Avg mu \cite{Glickman2013} (skill) & $0.65$ \\
    Avg phi (deviation) & $0.17$ \\
    Ties  & $1.90\%$ \\
    \bottomrule
    \end{tabular}%
  \label{tab:doda_stats}%
\end{table}%

To keep track of players’ skill levels, Dominion Online implements a rating system \cite{Meijer2017b} based on Glicko-2 \cite{Glickman2013}. Without going to deeply into the intricacies of the rating system, we suffice it to say that each player has an estimated skill $\mu$, and a deviation $\phi$, such that the system can be 95\% confident that the players skill is between $(\mu-2\phi)$ and $(\mu+2\phi)$. Larger increments/decrements of skill are applied when the skill levels of the players are further apart. The ratings and deviations for the players of each game are made available along with the game data. From Table \ref{tab:doda_stats}, we can see that on average, the players in our dataset have $\mu$ = .65 and $\phi$ = .17. For reference, the average player from our dataset has a 66\% chance of beating a player with $\mu$ = 0 and a 20\% chance of beating a top-10 player (using current ratings from the leaderboards). As a point of reference, the author has a $\mu$ of -0.07 at the time of this writing, and is the best Dominion player they know. Additionally, we present an excerpt of a game log in Table \ref{tab:game_log_exa} to familiarize the reader with the format of the data. 


\section{Benchmarks} \label{sec:benchmarks}

\setlength{\tabcolsep}{3pt}
\begin{table*}[ht!]
  \centering
  \caption{Play performance of Provincial and DQN-bot against heuristic baselines on 1000 games. The specific heuristic is expressed as a buy menu, which are explained in section \ref{sec:buy_menu}. In each game, the heuristic baselines are given the first turn.}
    \resizebox{\textwidth}{!}{\begin{tabular}{llrrrrrr}
    \toprule
          &       & \multicolumn{3}{c}{\textbf{Provincial}} & \multicolumn{3}{c}{\textbf{DQN-bot}} \\
    Bot   & Bot Buy Menu & \multicolumn{1}{c}{Win} & \multicolumn{1}{c}{Tie} & \multicolumn{1}{c}{Lose} & \multicolumn{1}{c}{Win} & \multicolumn{1}{c}{Tie} & \multicolumn{1}{c}{Lose} \\
    \midrule
    Big Money & (Gold, $99$), (Silver, $99$) & $972$   & $4$     & $24$    & $912$   & $27$    & $61$ \\
    Double Witch & (Witch, $1$), (Gold, $99$), (Witch, $1$), (Silver, $99$) & $596$   & $69$    & $335$   & $561$   & $63$    & $376$ \\
    Big Smithy & (Gold, $99$), (Smithy, $1$), (Silver, $99$) & $930$   & $7$     & $63$    & $851$   & $38$    & $111$ \\
    Big Militia & (Gold, $99$), (Militia, $1$), (Silver, $99$) & $946$   & $9$     & $45$    & $850$   & $31$    & $119$ \\
    Village/Smithy Engine\hphantom{a} & (Gold, $99$), (Smithy, $5$), (Militia, $1$), (Village, $5$), (Silver, $99$) & $994$   & $2$     & $4$     & $418$   & $24$    & $558$ \\
    Provincial & (Witch, $1$), (Gold, $99$), (Militia, $1$), (Witch, $1$), (Market, $3$), (Silver, $99$)\hphantom{a} & $550$   & $53$   & $397$   & $575$   & $62$    & $363$ \\
    \bottomrule
    \end{tabular}}%
  \label{tab:provincial_scores}%
\end{table*}%

To demonstrate that Dominion is ripe for reinforcement learning research, we’ve gathered and implemented some of the best bots currently available and compare them to a simple DQN bot. There are other bots that are competitive; however, are not suitable as baselines, and are outlined as such in \ref{DominionBots}. The simplest Dominion bots follow a set of easily articulable heuristics, while still regularly beating novice players. The most simple of these heuristics is known as “Big Money”. In essence, Big Money dictates the player to buy the most expensive money card available, unless they can afford a Province, in which case they should buy that. Many of the other common heuristic bots are simple variations on Big Money. For example, “Smithy Big Money” is the same as the original, but dictates that the player should buy a single Smithy when they are able to.

\subsection{Buy Menus} \label{sec:buy_menu}

In 2014, Fisher formalized this type of heuristic as a \textit{buy menu} \cite{Fisher2014}. This formalism is so useful that we were able to cast most of the best known heuristic bots in terms of a buy menu, which we list in Table \ref{tab:provincial_scores}. A buy menu is simply an ordered list of card/count tuples. To play a buy menu, a bot buys the leftmost card it can afford, decrementing the count of that card. Cards with zero counts are skipped over, ensuring no more than the specified number of each card is bought.  

\subsection{Provincial}

To answer the question of how to assemble the best buy menu given a card set, Fisher proposes the AI system Provincial \cite{Fisher2014}. This bot uses competitive coevolution, taking an arbitrary set of kingdom cards and evolving a list of leading strategies over the course of a few million games. The core of the bot’s strategy is a buy menu template with a fixed number of slots. During training, mutations can replace cards from the template, modify the purchase count, swap the order of cards, and change the victory card purchase thresholds.

We test Provincial against some of the best known heuristics (also cast as buy menus) and show results in Table \ref{tab:provincial_scores}; Provincial clearly sweeps the board. This is despite the fact that the heuristic baselines always went first, which seems to carry a significant advantage. This is evidenced by the fact that in the Provincial-Provincial matchup, the first player won 55\% of the time, whereas the second player only won 40\% of the time. In addition to this strong performance, Provincial’s relatively simple strategy representation (the buy menu) allows it to provide a very clear and explanative visual representation of the leading strategies, as pictured in Fig. \ref{fig:main}(b). This property further increases the usefulness of Provincial as a benchmark, as it allows a researcher to quickly identify leading strategies on a new card set they wish to test their agents on.

\subsection{DQN-bot}

We implement a simple RL-based buyer bot using Rainbow DQN \cite{Hessel2018}, trained to play Dominion via self-play. Our intention with this bot is not to break new ground in the field of RL but to show that recent developments in RL can easily outpace the current state of the art in Dominion. For that reason, we present here only necessary detail and encourage the reader to refer to our code for the minutiae. The bot is a small fully connected neural network (with two hidden layers, both size 256) that takes in the game state representation and action mask, and returns which card to gain. The bot only makes gain decisions; action card decisions are handled by the same heuristics that Provincial uses. We train for 1000 steps, which comes out to about 7000 games of Dominion and 45 minutes of training time on an NVIDIA 1080 Ti GPU. This is comparable to how long Provincial takes to train with 32 generations on a modern computer. 


We see in Table \ref{tab:provincial_scores} that our DQN-bot beats or ties Provincial almost two thirds of the time. It also clearly beats all of the heuristics except for the engine, against which it struggles. We hypothesize that this is because the learner did not explore this particular combo during its training and so got confused when it saw its opponent take this strategy. Indeed when we examined the game logs, the bot was behaving strangely, buying Duchies too early, and even buying Curses. We believe this problem is solvable, likely with a combination of more training and a more sophisticated game representation, but we leave that to future work. In particular, we believe that this simple instance of a DQN-bot shows that existing RL based methods can crush existing bots.

\section{Conclusion}

We started off by discussing some of the properties of Dominion that make it particularly well suited for AI research. Specifically, these include its huge variety of card combinations, a vibrant player community, and various affordances for making the state space more or less complex. Among these affordances are decisions of different complexity/importance, and a gradient of card complexity, which both serve to lower the barrier to entry for researchers trying out new ideas. Next we explored the Dominion Online Dataset, which consists of 30,000 high quality base game records -- and up to 2 million total games. These games were played by experienced Dominion players, and can serve as a strong pretraining tool for RL based AI approaches. Finally, we explored the current state of the art in Dominion playing bots, and implemented a number of heuristic based bots, along with the genetically evolved bot Provincial. We compared these to a simple RL-based bot using recent innovations in RL, and showed that the RL-based bot generally outperformed the existing baselines. 


\clearpage
\acknowledgments{We would like to thank Ben Zhang for helping us understand and extend the DomRL codebase, ceviri for giving us free server time and assistance with downloading games from Dominion, Markus for his help with extracting the base game ids, and Dan Brooks for his good advice.}


\bibliography{example}  

\begin{thebibliography}{13}
\providecommand{\natexlab}[1]{#1}
\providecommand{\url}[1]{\texttt{#1}}
\expandafter\ifx\csname urlstyle\endcsname\relax
  \providecommand{\doi}[1]{doi: #1}\else
  \providecommand{\doi}{doi: \begingroup \urlstyle{rm}\Url}\fi

\bibitem[Vaccarino(2008)]{Vaccarino2008}
D.~X. Vaccarino.
\newblock Dominion, 2008.
\newblock URL \url{http://riograndegames.com/Game/ 278-Dominion}.

\bibitem[Gam()]{GameWiki}
Dominion strategy wiki.
\newblock URL \url{http://wiki.dominionstrategy.com/}.

\bibitem[Meijer(2017)]{Meijer2017}
S.~Meijer.
\newblock Dominion online, 2017.
\newblock URL \url{dominion.games}.

\bibitem[Glickman(2013)]{Glickman2013}
M.~E. Glickman.
\newblock Example of the glicko-2 system.
\newblock Technical report, Boston University, 11 2013.
\newblock unpublished.

\bibitem[Meijer(2017)]{Meijer2017b}
S.~Meijer.
\newblock Rating details, 3 2017.
\newblock URL \url{http://forum.shuffleit.nl/index.php?topic=1679.msg5891}.

\bibitem[Fisher(2014)]{Fisher2014}
M.~D. Fisher.
\newblock Provincial: A kingdom-adaptive ai for dominion.
\newblock Technical report, Stanford, 2014.
\newblock unpublished.

\bibitem[Hessel et~al.(2018)Hessel, Modayil, Van~Hasselt, Schaul, Ostrovski,
  Dabney, Horgan, Piot, Azar, and Silver]{Hessel2018}
M.~Hessel, J.~Modayil, H.~Van~Hasselt, T.~Schaul, G.~Ostrovski, W.~Dabney,
  D.~Horgan, B.~Piot, M.~Azar, and D.~Silver.
\newblock Rainbow: Combining improvements in deep reinforcement learning.
\newblock In \emph{Proceedings of the AAAI Conference on Artificial
  Intelligence}, volume~32, 2018.

\bibitem[van~der Heijden(2014)]{Heijden2014}
R.~R. van~der Heijden.
\newblock An analysis of dominion.
\newblock Master's thesis, Leiden University, 8 2014.

\bibitem[Fynbo and Nellemann(2010)]{Fynbo2010}
R.~B. Fynbo and C.~S. Nellemann.
\newblock Developing an agent for dominion using modern ai-approaches.
\newblock Master's thesis, IT-University of Copenhagen, 2010.

\bibitem[Mahlmann et~al.(2012)Mahlmann, Togelius, and Yannakakis]{Mahlmann2012}
T.~Mahlmann, J.~Togelius, and G.~N. Yannakakis.
\newblock Evolving card sets towards balancing dominion.
\newblock In \emph{2012 IEEE Congress on Evolutionary Computation}, pages 1--8.
  IEEE, 2012.

\bibitem[Jansen and Tollisen(2014)]{Jansen2014}
J.~V. Jansen and R.~Tollisen.
\newblock \emph{An AI for Dominion Based on Monte-Carlo Methods}.
\newblock PhD thesis, University of Agder, 6 2014.

\bibitem[Winder(2014)]{Winder2014}
R.~K. Winder.
\newblock Methods for approximating value functions for the dominion card game.
\newblock \emph{Evolutionary Intelligence}, 6\penalty0 (4):\penalty0 195--204,
  2014.

\bibitem[Fenner(2014)]{Fenner2014}
C.~Fenner.
\newblock Ann assisted heuristic tree search in computer dominion.
\newblock Master's thesis, University of Oklahoma, 2014.

\end{thebibliography}

\newpage
\appendix
\section{Appendix}
\subsection{Game Logs}
We present an excerpt of a game log from the Dominion Online Dataset in Table \ref{tab:game_log_exa} to familiarize the reader with the format of the data. 
\begin{table}[ht!]
  \centering
  \caption{Excerpt of a game log from the Dominion Online Dataset showing two full turns of gameplay. Each game log lists the ids of the two players, followed by a list of the supply pile cards, and the game id. Then each line of the log corresponds to some event that the game client processed, in the order the players caused those events. For example, the card Harbinger first draws the player a card, then increases the number of actions they can take this turn by 1, and finally allows them to look through their discard pile and optionally topdeck a card. This sequence of events is reflected by lines 149 - 153. }
    \begin{tabular}{p{0.95\linewidth}}
    \verb+<+player\_id1\verb+>+$\sim$\verb+<+player\_id2\verb+>+ \\
    Supply: 10 Curse, 60 Copper, 40 Silver, 30 Gold, 14 Estate, 8 Duchy, 8 Province, 10 Bandit, 10 Bureaucrat, 10 Cellar, 10 Harbinger, 10 Militia, 10 Poacher, 10 Remodel, 10 Throne Room, 10 Village, 10 Workshop \\
    0:P0 - GAME\_META\_INFO (\verb+<+game\_id\verb+>+): \\
    1:P1 - STARTS\_WITH: 7 Copper \\
    2:P1 - STARTS\_WITH: 3 Estate \\
    … \\
    148:P1 - NEW\_TURN (10, 0): \\
    149:P1 - PLAY: 1 Harbinger \\
    150:P1 - DRAW: 1 Cellar \\
    151:P1 - GETS\_ACTION (1): \\
    152:P1 - LOOK\_AT: 3 Copper, 1 Province, 1 Harbinger, 1 Militia, 1 Gold \\
    153:P1 - TOPDECK: 1 Gold \\
    154:P1 - PLAY: 1 Poacher \\
    155:P1 - DRAW: 1 Gold \\
    156:P1 - GETS\_ACTION (1): \\
    157:P1 - GETS\_COIN (1): \\
    158:P1 - PLAY: 1 Cellar \\
    159:P1 - GETS\_ACTION (1): \\
    160:P1 - DISCARD: 2 Estate \\
    161:P1 - DRAW: 1 Copper, 1 Remodel \\
    162:P1 - PLAY: 1 Bandit \\
    163:P1 - "GAIN: 1 Gold \\
    164:P2 - REVEAL: 2 Village \\
    165:P2 - DISCARD: 2 Village \\
    166:P1 - PLAY\_TREASURES\_FOR (4): 1 Copper, 1 Gold \\
    167:P1 - BUY\_AND\_GAIN: 1 Throne Room \\
    168:P1 - SHUFFLES \\
    169:P1 - DRAW: 2 Copper, 2 Estate, 1 Bandit \\
    170:P2 - NEW\_TURN (10, 0): \\
    171:P2 - PLAY: 1 Bandit \\
    172:P2 - "GAIN: 1 Gold \\
    173:P1 - REVEAL: 1 Estate, 1 Throne Room \\
    174:P1 - DISCARD: 1 Estate, 1 Throne Room \\
    175:P2 - PLAY\_TREASURES\_FOR (3): 3 Copper \\
    176:P2 - BUY\_AND\_GAIN: 1 Village \\
    177:P2 - SHUFFLES \\
    178:P2 - DRAW: 1 Copper, 1 Gold, 1 Estate, 1 Bandit, 1 Throne Room \\
    … \\
    \end{tabular}%
  \label{tab:game_log_exa}%
\end{table}%

\subsection{Additional Dominion Agents}
\label{DominionBots}
We focus this section on existing Dominion playing AIs, specifically the ones which we did not implement, and explain why we believe they are not suitable benchmarks.

Van der Heijden \cite{Heijden2014} develops a framework of definitions for reasoning about deck building games, focusing their analysis on the equivalency of different game states, and choosing the optimal buy decision in a certain state. However, they make assumptions that overly simplify and rigidify the game, rendering their AIs unusable to us. 

Fynbo and Nellemann \cite{Fynbo2010} develop a bot consisting of three parts: a neural network to estimate the number of turns remaining in a game, and two NEAT-powered networks to take actions and make buys. Their bot beat Big Money (a common buy-heuristic explained in section \ref{sec:benchmarks}) 55\% of the time and a finite state machine bot 48\% of the time; in section \ref{sec:benchmarks} we review the performance of Provincial and our DQN-bot against Big Money and other heuristic bots and show that this is not very impressive. Mahlmann, Togelius, and Yannakakis \cite{Mahlmann2012} use Dominion as a complex test-bed for their genetic search algorithm whose purpose is to aid game designers in coming up with more balanced rulesets. The authors implement a number of quite complex rule based agents that they use to test this algorithm. However, these bots demonstrate mixed performance when playing against Fynbo and Nellemann’s AI, and do not seem strong enough in general to warrant reimplementing in our environment.

In addition to performance, we were looking for existing bots that had the potential to scale to the rest of the Dominion expansions beside the base game. Jansen and Tollisen \cite{Jansen2014} build an AI for Dominion using Monte Carlo methods that handily beat moderately-experienced human players and do quite well against heuristic baselines. However, for their approach they limit all their tests to just 10 of the simplest action cards from the base game. Although they suggest that their approach could be scalable to the complete Dominion, it is not at all trivial to do so, and we considered it outside the scope of this work. Winder’s \cite{Winder2014} method uses a neural network to predict a value approximation of every possible next game state. This approach is not scalable for two reasons: their genetic training process is quickly overwhelmed by the totality of Dominion cards, and the value approximation requires an accurate prediction of the next state, which is not always possible. Fenner \cite{Fenner2014} applies a hybrid approach, combining heuristics, neural networks, and tree search to build a Dominion AI. The most interesting aspect of the approach is that it vectorizes cards by their effects. Because some cards reuse effects from earlier cards, the intent of this representation is to possibly generalize to previously unseen cards whose effects are a subset of previously seen effects. Although the specifics of the approach are a bit outdated and costly to scale, we believe this idea could make for interesting future work, perhaps using something like an autoencoder instead of hand engineering.

\end{document}